\pdfoutput=1
\documentclass[11pt]{article}

\usepackage[]{EMNLP2023}

\usepackage{times}
\usepackage{latexsym}
\usepackage{graphicx}
\usepackage{hyperref}
\usepackage{multirow}
\usepackage[ruled,linesnumbered]{algorithm2e}
\usepackage{booktabs}

\usepackage[T1]{fontenc}
\usepackage[utf8]{inputenc}

\usepackage{microtype}

\usepackage{inconsolata}

\title{From Static to Dynamic: A Continual Learning Framework for Large Language Models}

\author{
    Mingzhe Du $^{1,2}$ \and 
    Anh Tuan Luu $^1$ \and
    Bin Ji $^2$ \and 
    See-Kiong Ng $^2$ \\
    $^{1}$Nanyang Technological University \\ 
    $^{2}$National University of Singapore \\ 
    {\tt \{mingzhe001,anhtuan.luu\}@ntu.edu.sg; \{jibin,seekiong\}@nus.edu.sg}
}

\begin{document}
\maketitle
\begin{abstract}
The vast number of parameters in large language models~(LLMs) endows them with remarkable capabilities, allowing them to excel in a variety of natural language processing tasks. However, this complexity also presents challenges, making LLMs difficult to train and inhibiting their ability to continuously assimilate new knowledge, which may lead to inaccuracies in their outputs. To mitigate these issues, this paper presents \textbf{DynaMind}, a novel continual learning framework designed for LLMs. DynaMind incorporates memory mechanisms to assimilate new knowledge and modular operators to enhance the model inference process with the newly assimilated knowledge, consequently improving the accuracies of LLMs' outputs. Benchmark experiments demonstrate DynaMind's effectiveness in overcoming these challenges. The code and demo of DynaMind are available on GitHub: \url{https://github.com/Elfsong/DynaMind}.

\end{abstract}

\section{Introduction}
The advent of large language models (LLMs) marks a revolutionary shift in the realm of artificial intelligence, showcasing extraordinary competence across a broad array of tasks~\citep{brown2020language, OpenAI_2023, touvron2023llama, penedo2023refinedweb}. With their capacity to generate human-like texts and address intricate inquiries, LLMs have fundamentally redefined our understanding of machine learning capabilities. However, like any technological breakthrough, LLMs come with their own set of limitations. 

One of the primary drawbacks lies in their static nature~\citep{kemker2018measuring}. The knowledge embedded within LLMs is confined to the fixed parameters established during the training phase. As a result, the process of integrating new knowledge to LLMs through fine-tuning is not only computationally demanding but also prone to catastrophic forgetting~\citep{dong2021should,scao2022language}. Moreover, as the knowledge is implicitly encoded within the parameter space, LLMs lack the ability to articulate their understanding explicitly, which may result in the model generating unfounded or hallucinated information, thereby compromising the reliability of their outputs~\citep{azamfirei2023hallucinations}. These inherent challenges significantly impede the capacity of LLMs to assimilate new knowledge and adapt to evolving environments. 

To tackle the above issues, we propose DynaMind, a novel continual learning framework for LLMs. Continual learning encapsulates the capacity to dynamically adapt cognition by assimilating new knowledge from the environment over time, acting as a fundamental manifestation of human intelligence~\citep{hadsell2020embracing}. 
%This capability is vital for LLMs to uphold their credibility and accuracy, particularly in a world where information is perpetually evolving. 
Motivated by human continual learning \citep{eichenbaum2004hippocampus, hadsell2020embracing}, we carefully design an independent memory module in DynaMind, which empowers LLMs to heuristically search knowledge, store knowledge in the memory and recall relevant knowledge from the memory in subsequent inference. This mechanism equips LLMs with continual learning ability without tuning any model parameters.

To evaluate DynaMind's continual learning capabilities, we conducted extensive experiments centered on three aspects: \emph{Knowledge-driven Complex Reasoning}, \emph{Knowledge Credibility Perception}, and \emph{Knowledge Manipulation}. Detailed definitions and analyses of each aspect are presented in Section~\ref{sec:evaluation_benchmark}. The empirical results corroborated that DynaMind substantially augments the continual learning capacity of LLMs, paving the way for more dynamic and adaptable AI systems.

\section{Related Work}
\subsection{AutoGPT}
AutoGPT is an experimental open-source application to energize the advanced capabilities of GPT models. It is designed to autonomously iterate the "Chain-of-Thought" in pursuit of a specified objective~\citep{autogpt}. Unlike interactive systems such as ChatGPT, which {necessitates} manual input for each iteration, AutoGPT is conceived to independently manage the task of achieving broader objectives without human intervention. As one of the first fully self-contained LLM instances, AutoGPT effectively pushes the boundary of where AI can achieve. However, the practicality of AutoGPT is still greatly limited when confronted with complex scenarios due to its lack of long-term memory and task coordination capabilities.

\subsection{BabyAGI}
BabyAGI is an AI task management system comprising LLMs and vector search APIs, which synergizes for task prioritization and execution~\citep{babyagi}. The main idea behind this system is to recursively create new tasks based on the results of previous tasks and predefined objectives and then systematically solve them in a step-wise manner.
During this phase, BabyAGI also stores and retrieves intermediate results in its memory to augment the inference context, while a potential shortcoming is that BabyAGI lacks the capability to autonomously discern and refresh obsolete knowledge stored in its memory. This drawback of referencing outdated knowledge may be progressively magnified along with the inference process, thereby impacting the final results. 

% \subsection{Langchain}
% Langchain is a development toolkit painstakingly designed to meet the exact needs of large language model applications~\citep{Chase_LangChain_2022}. Equipped with a comprehensive array of tools and resources, this toolkit supports developers involved in LLM-related projects by offering utilities such as prompt template management, streamlined invocation of language models, and vector retrieval libraries. DynaMind intensively utilized the Langchain library as well.

\section{DynaMind}
\begin{figure*}[ht]
    \centering
    \includegraphics[width=\textwidth]{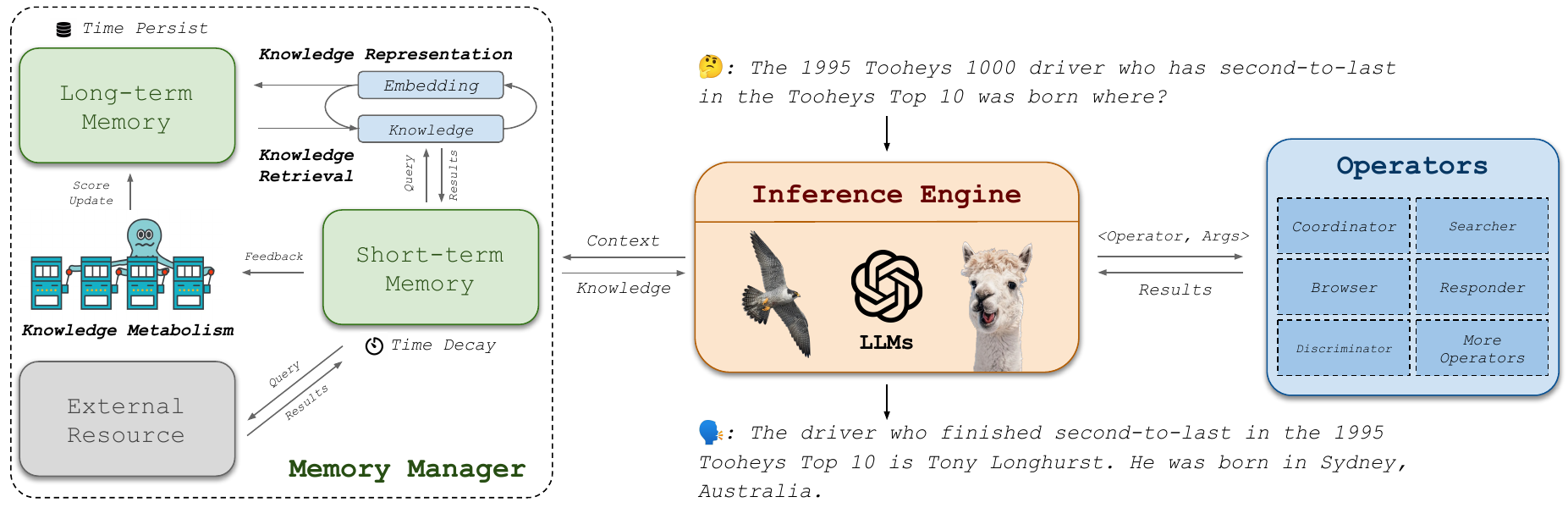}
    \caption{The system overview of DynaMind.}
    \vspace*{-1.5em}
    \label{fig:system_overview}
\end{figure*}

In this section, we will delve into the architecture and workflow of DynaMind. As illustrated in Figure~\ref{fig:system_overview}, the system is primarily composed of three main components: \emph{Inference Engine}, \emph{Memory Manager}, and \emph{Operators}. In a nutshell, the intellectual core, \emph{Inference Engine} retrieves the knowledge required for decision-making from \emph{Memory Manager} based on contexts and then generates subsequent instructions, which will be executed by \emph{Operators}. 

% By engaging in the continuous process of knowledge retrieval, DynaMind actively enhances the credibility of effective knowledge while gradually diminishing the credibility of outdated information until it is eventually disregarded. This mechanism enables a dynamic knowledge metabolism within the system, ensuring that the most reliable and up-to-date information is consistently prioritized.

\subsection{Inference Engine}
\emph{Inference Engine} is powered by one or more LLMs which possess exceptional natural language reasoning capabilities. By integrating contextual information into carefully crafted prompt templates, DynaMind empowers these LLMs to produce structured instructions readily interpretable by \emph{Operators}. Formally, \emph{Inference Engine} can be formalized as follows:

\[IE(O_{in}, P_{in}) \rightarrow List[(O_{out}, P_{out})]\]

{Here,} $ IE $ stands for \emph{Inference Engine}, $ O_{in} $ denotes the running operator, and $ P_{in} $ represents the input parameters associated with the running operator. The output of \emph{Inference Engine} is represented as a list of subsequent command tuples, denoted by $ List[(O_{out}, P_{out})] $. 
Each tuple within the list consists of the output operator $O_{out}$ and its corresponding parameters $P_{out}$. It is important to note that the output list can be empty, both $ O_{in} $ and $ O_{out} $ should be constrained within the operator list, and both $ P_{in} $ and $ P_{out} $ should be correctly parsed by the corresponding operator. Moreover, \emph{Inference Engine} internally employs a FIFO~(First In, First Out) priority queue to keep track of the operators that require execution $ Queue[(O_{in}, P_{in})] $. If the output is an empty list, i.e., $List[None]$, the engine pops the current operator directly. Otherwise, the engine replaces the current operator with the output list $ List[(O_{out}, P_{out})] $ and proceeds with the traversal until the queue becomes empty. This recursive solution method can gradually decompose the original query into sub-problems when the system cannot solve the query directly. To handle the potential infinite recursive problem, DynaMind allows users to set a maximum recursion depth, thereby controlling the extent of problem decomposition. 
% Once the maximum recursion depth is reached, the system will promptly return the first command in the command queue to the user, seeking their assistance in solving the problem.

\begin{figure}
    \centering
    \includegraphics[width=\linewidth]{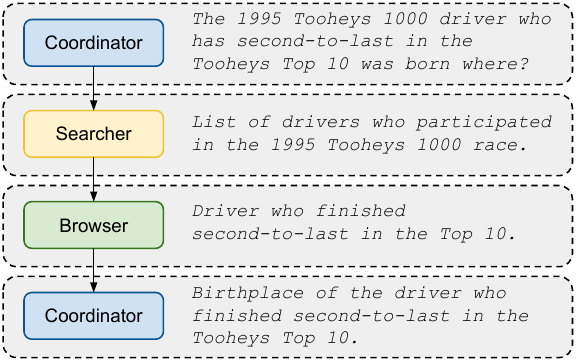}
    \caption{The pipeline of example "\textit{The 1995 Tooheys 1000 driver who has second-to-last in the Tooheys Top 10 was born where?}".}
    \vspace*{-1.5em}
    \label{fig:pipeline_example}
\end{figure}

For instance, let's consider the query: "\textit{The 1995 Tooheys 1000 driver who has second-to-last in the Tooheys Top 10 was born where?}". The first item in the operator queue will be ("Coordinate", {"query": "\textit{The 1995 Tooheys 1000 driver who has second-to-last in the Tooheys Top 10 was born where?}"}). The desired output should resemble the case shown in Figure~\ref{fig:pipeline_example}.

In light of differences in model parameters and training data, the capabilities of LLMs can differ significantly. Therefore, different LLMs can be employed for various tasks during the inference process. For example, while relatively small-scale LLMs may suffice for text summarization, large-scale LLMs are required for logical reasoning. Rather than relying on a single model, carefully selecting the appropriate model for each task not only enhances inference speeds but also improves overall performance. 
% To maintain consistency in utilizing diverse models, we have developed a collection of compatible interfaces. With these interfaces, DynaMind can seamlessly employ different models by simply defining the relevant prompt templates and accessing them through a unified interface. The details regarding operators and prompt templates will be discussed in Section~\ref{sec:operators}. 
Currently, DynaMind supports the following LLMs: OpenAI GPT-3.5 and GPT-4\footnote{\url{https://openai.com/blog/openai-api}}, Llama series models\footnote{\url{https://huggingface.co/huggyllama/llama-30b}}, and Falcon series models\footnote{\url{https://huggingface.co/tiiuae/falcon-40b}}.

\subsection{Memory Manager}
\emph{Memory Manager} plays a pivotal role in storing and organizing the memories of DynaMind. It consists of five interconnected modules: Knowledge Representation, Knowledge Retrieval, Long-term Memory, Short-term Memory, and Knowledge Metabolism. \emph{Memory Manager} works hand in hand with \emph{Inference Engine}, engaging in frequent interactions to ensure a continuous acquisition and updating of the knowledge necessary for inferencing. This section provides an overview of the functionalities within \emph{Memory Manager} and elaborates on their interactions with \emph{Inference Engine}.

\textbf{Knowledge Representation} is responsible for encoding knowledge in a format that can be efficiently processed by DynaMind. Typically, there are several common methods used for knowledge representation, such as Bag of Words (BOW)~\citep{zhang2010bow},Word2Vec~\citep{word2vec}, knowledge graph ~\citep{luu2014taxonomy,luu2016learning}, or fine-tuning a Transformer encoder \citep{vaswani2017attention}. To uphold the close alignment with the LLMs' semantic space, we utilized the embeddings generated by LLMs as the knowledge representation. For OpenAI series models, we can directly retrieve the embedding through their official APIs\footnote{\url{https://platform.openai.com/}}. For other open-source models, we choose the last hidden state as the representation.

\textbf{Knowledge Retrieval} utilizes vector search to construct an index on DynaMind's memory, enabling the rapid identification and retrieval of the most relevant knowledge. By leveraging the Langchain library~\citep{Chase_LangChain_2022}, DynaMind can effortlessly access and shift between distinct vector indexing schemes. This module evaluates the relevance of knowledge within the given context, empowering DynaMind to access pertinent knowledge throughout the reasoning process efficiently.

In DynaMind's memory, every piece of knowledge is represented as a triple <Context, Key, Value>. The "Context" component houses the contextual data relating to the knowledge. The "Key" signifies the vectorized representation of the "Context", while the "Value" embodies the specific substance of the knowledge.

\textbf{Long-term Memory} acts as a permanent knowledge repository within DynaMind, accumulating and retaining large amounts of information over time. The repository holds a collection of various knowledge from a variety of sources, such as previous interactions, external databases, and the Internet. DynaMind harnesses the power of long-term memory to enhance its reasoning ability by integrating past experience and acquired knowledge. Furthermore, users can actively manipulate the knowledge engaged in the inference by explicitly updating the long-term memory, thereby granting them heightened control over DynaMind's cognitive processes.

\textbf{Short-term Memory} functions as a temporary workspace for \emph{Inference Engine}. To incorporate external information into the inference process,  all the knowledge stored in short-term memory is consolidated with the user query and provided to \emph{Inference Engine}. However, due to the context length limitation of LLMs, short-term memory primarily retains the immediate knowledge required to support the current reasoning task. To ensure efficient use of short-term memory, each piece of knowledge added is assigned a variable that gradually decreases over time. Once this variable falls below a predefined threshold, the knowledge will be popped from the short-term memory, unless it is recalled again. This mechanism mimics human cognitive behavior during reasoning tasks. It enables DynaMind to adapt to dynamic situations swiftly by selectively retrieving and discarding pertinent knowledge in short-term memory.

\textbf{Knowledge Metabolism}
The knowledge stored in long-term memory may expire as time progresses. For instance, the statement "Donald Trump is the current President of the United States" was accurate in 2020, but it became outdated after 2021. This issue is prevalent in real life but discovering and renovating such knowledge necessitates a painstaking endeavor~\citep{jiang2016timekg}. To alleviate this predicament, DynaMind is designed to adaptively modulate the credibility of long-term memory knowledge in response to varying contextual conditions. We called this mechanism \emph{Knowledge Metabolism}.

Inspired by the Multi-Armed Bandit (MAB) algorithm, we abstract this issue as an exploration-exploitation problem~\citep{zeng2016onlinemab, auer2002finitemab}. During the cold start period, \emph{Memory Manager} retrieves $ K_t $ knowledge pieces. At this stage, the model lacks prior knowledge about the credibility associated with each knowledge piece $ k $. To figure out the most reliable knowledge for the given context $ c $, we assume a linear relationship between the credibility and the context. Consequently, the problem is transformed into a MAB estimation task, wherein we employ contextual features $ v_{t,k} $ of combining both $ k $ and $ c $ to estimate the credibility and its confidence interval. Subsequently, we choose the knowledge piece with the highest upper limit in its confidence interval $ p_{t,k} $. This approach emphasizes exploration when the confidence interval for $ k $ is wide, indicating fewer selections and increased uncertainty, representing the risky component of the algorithm. Conversely, if the confidence interval for $ k $ is narrow, implying an increased certainty regarding their credibility, the algorithm tends to favor knowledge pieces with higher means, reflecting the conservative and cautious side of the algorithm. Algorithm~\ref{alg:knowledge_metabolism} illustrates the specific process of the algorithm.

\vspace{-0.5em}
\begin{algorithm}
    \SetAlgoLined
    \KwIn{$\alpha \in \mathbf{R}^+ $}
    \For{t = 1,2,3,...,T} {
        Contextual features: $ k \in K_t; v_{t,k} \in \mathbf{R}^d $ \\
        \For{all $ k \in K_t $} {
            \eIf {$ k $ is new} {
                $ A_k \leftarrow I_{d \times d} $ \\
                $ b_k \leftarrow 0_{d \times 1} $ \\
            } {
                $ \theta_k \leftarrow A_{k}^{-1} b_{k} $ \\
                $ p_{t,k} \leftarrow \theta^{\mathrm{T}}_k v_{t,k} + \alpha \sqrt{x^{\mathrm{T}}_{t,k} A^{-1}_{k} x_{t,k}} $ \\
            }
        }
        Choose arm $ k_o = \arg \max_{k \in K_t} p_{t,k} $, and observe a real-valued payoff $ r_t $ \\
        $ A_{k_o} \leftarrow A_{k_o} + x_{t, k_o} x^{\mathrm{T}}_{t, k_o} $ \\
        $ b_{k_o} \leftarrow b_{k_o} + r_t x_{t, k_o} $ \\
    }
    \caption{Knowledge Metabolism}
    \label{alg:knowledge_metabolism}
\end{algorithm}

\vspace{-1.5em}
\subsection{Operators}
\label{sec:operators}
DynaMind provides a set of atomic operators that harness the power of an LLM with well-designed prompt templates. By permuting these essential operators, DynaMind is able to break down intricate problems into smaller, manageable sub-problems recursively, enabling step-by-step resolution. Moreover, the operators incorporate integrated functionalities such as invoking search engines and browsing websites. Table~\ref{tab:operators} shows the definition of the built-in operators. For specific operator formats and illustrative examples, please refer to Appendix Table~\ref{tab:operators_details}.

\begin{table*}
\centering
\renewcommand\tabcolsep{12.2pt}
\begin{tabular}{ll}
\toprule
\textbf{Name} & \textbf{Definition}                                                                                                    \\ \midrule
Coordinator   & \textit{Context: List{[}Knowledge: str{]}, Query: str → List{[}(Operator: str, Args: dict){]}} \\
Searcher      & \textit{Query: str → Results: List{[}Result: dict{]}}                                                                  \\
Browser       & \textit{Path: str, Query: str → Answer: str}                                                                     \\
Responder     & \textit{Context: List{[}Knowledge: str{]}, Query: str → Response: str}                                                 \\
Discriminator & \textit{Context: List{[}Knowledge: str{]}, Query: str, Response: str → Validity: boolean}                              \\ \bottomrule
\end{tabular}
\caption{The definition of DynaMind operators.}
\vspace*{-1.5em}
\label{tab:operators}
\end{table*}

\textbf{Coordinator} plays a vital role in efficiently managing and coordinating the activities of other operators. As central control operators, they hold the responsibility of recursively breaking down queries into sub-problems based on the context. This breakdown continues until the sub-problems can be handled independently by other operators, enabling efficient collaboration among them to solve complex problems effectively. 

\textbf{Searcher} in DynaMind provides both keyword-based and vector-based searches to supply context-related knowledge for \emph{Inference Engine}. By leveraging the keyword-based search, DynaMind offers a set of SQL-like interfaces that enable a fine-grained level of control over the data stored in memory. On the other hand, vector search can capture implicit relationships between context and query, granting the system greater flexibility in retrieving knowledge. As a result, the hybrid search solution allows DynaMind to retrieve knowledge in a more nuanced and accurate manner, taking into account the underlying meaning and connections between different pieces of knowledge.

\textbf{Browser} runs as a cohesive operator for reading file resources. Currently, it supports parsing HTML/PDF files and allows to support more extensions through customizing the operator. To facilitate inference by \emph{Inference Engine}, DynaMind handles files whose content exceeds the token limit of LLMs in the following manner: 1) It employs the boilerplate removal library~\citep{leonhardt2020boilerplate} to eliminate irrelevant content from the files. 2) It utilizes text summarization techniques to compress the file content into an appropriate length that can be processed effectively.

\textbf{Responder} is invoked to consolidate all knowledge in the short-term memory and generate a response if the coordinator operator determines that the current context and acquired knowledge are sufficient to provide an answer to the user's query. It is important to note that the response does not immediately deliver to the user. Instead, it calls upon the discriminator operator to evaluate whether the generated result fulfills the user's requirements. Should the response satisfy the user's expectations, it is subsequently relayed back to the user. Otherwise, DynaMind reserves the unsatisfactory response in its short-term memory to avoid replicating a similar case, and then activates the coordinator to generate a more suitable response.

\textbf{Discriminator} evaluates whether the response meets the user's requirements. If the response is satisfactory, the discriminator calculates the contribution of each piece of knowledge in the current short-term memory to the response and increases the credibility of the corresponding knowledge proportionally. On the contrary, if the response does not meet the requirements, the discriminator accordingly lowers their credibility. Knowledge with high credibility will receive more attention in the retrieval phase, while knowledge with low credibility will be ignored.

\section{Evaluations and Benchmarks}
\label{sec:evaluation_benchmark}
To assess the effectiveness of DynaMind in the context of continual learning, we devised three benchmark datasets: \emph{Knowledge-driven Complex Reasoning}, \emph{Knowledge Credibility Perception}, and \emph{Knowledge Manipulation}. Subsequently, we conducted a series of quantitative analyses on these datasets, comparing the performance of a variety of LLMs within the DynaMind framework across these three distinct datasets

\subsection{Knowledge-driven Complex Reasoning}
The hallucinations that frequently arise in LLMs during complex reasoning tasks present a significant challenge in achieving convincing results~\citep{azamfirei2023hallucinations}. Additionally, these models often struggle to provide a comprehensive and accurate reasoning process~\cite{wang2022self}. In light of these issues, our objective with DynaMind is to incorporate memory capabilities into LLMs, enabling it to store the reasoning process and relevant knowledge in short-term memory. By doing so, DynaMind can employ heuristic strategies to break down complex problems into more manageable sub-problems and retain the solutions to these sub-problems in its short-term memory. Subsequently, it integrates the results from all the sub-problems to derive a conclusive outcome supported by a complete reasoning chain. Furthermore, by recalling pertinent knowledge from memory during the sub-problem reasoning process, DynaMind effectively mitigates the potential for hallucinations caused by insufficient knowledge in LLMs. 

We selected a set of 200 high-quality complex knowledge-driven reasoning question-answer pairs from the ComplexWebQA dataset~\citep{talmor2018web}. Table~\ref{tab:knowledge_reasoning} reports the performance of different LLMs on this dataset. In the situation without external resources, Falcon achieved the best results with an accuracy of 17.5~\%, whereas OpenAI GPT-4 substantially improved the accuracy to 92.5~\% when utilizing the power of DynaMind.

\begin{table}
\centering
\renewcommand\tabcolsep{11pt}
\begin{tabular}{lcc}
\toprule
\textbf{Model Name} & \textbf{Basic}    & \textbf{DynaMind}                          \\ 
\midrule
OpenAI GPT-3.5      &\ \ 8.5            & 89.0 \scriptsize{(+80.5)}                  \\
OpenAI GPT-4        & 16.0              & \underline{92.5} \scriptsize{(+76.5)}      \\
Falcon-40B          & \underline{17.5}  & 85.0 \scriptsize{(+67.5)}                  \\
Llama-30B           &\ \ 6.0            & 56.5 \scriptsize{(+50.5)}                  \\ 
\bottomrule
\end{tabular}
\caption{Accuracy on Knowledge-driven Reasoning.}
\vspace*{-1em}
\label{tab:knowledge_reasoning}
\end{table}

\subsection{Knowledge Credibility Perception}
The knowledge retained in long-term memory may become conflicting or obsolete over time. Identifying and upgrading such knowledge manually can be a strenuous and time-consuming process~\citep{jiang2016timekg}. Consequently, we conceived a task to determine how DynaMind could automatically assess the trustworthiness of each piece of information in long-term memory within various contexts.

To quantitatively gauge this capability, we selected 200 unique statements from the SciFact dataset~\citep{wadden2020fact}. Initially, we generated a counterfactual statement for each original statement using OpenAI GPT-4. Both of these statements were stored in long-term memory with identical initial credit scores. We then converted the original statements into question-answer pairs and prompted DynaMind to respond to these questions. After several cycles of the knowledge metabolism process, the credibility of the original statements is expected to increase, while the credibility of the constructed statements should correspondingly decay. Table~\ref{tab:knowledge_perception} depicts the final ratio of original knowledge in the long-term memory after various epochs of knowledge metabolism, where the credibility of the original statement surpasses that of the counterfactual ones. It's worth noting that Falcon-40B demonstrated comparable performance to GPT-4 but with a four times smaller model size.

\begin{table}
\centering
\renewcommand\tabcolsep{11pt}
\begin{tabular}{lcc}
\toprule
\textbf{Model Name} & \textbf{3 epochs}     & \textbf{5 epochs}     \\ \midrule
OpenAI GPT-3.5      & 48.5                  & 61.5                  \\
OpenAI GPT-4        & \underline{79.0}      & \underline{84.5}      \\
Falcon-40B          & 72.5                  & 78.0                  \\
Llama-30B           & 23.0                  & 42.5                  \\ \bottomrule
\end{tabular}
\caption{Performance on Credibility Perception.}
\vspace*{-1em}
\label{tab:knowledge_perception}
\end{table}

\subsection{Knowledge Manipulation}
DynaMind equips language models with the ability to assimilate new knowledge without the modification of their parameters, which ensures models are adaptive amidst a perpetually evolving environment. In this section, we evaluated the capability of DynaMind to handle knowledge from three aspects: creation, update, and deletion.

For the creation aspect, we extracted 200 news as knowledge from the latest WikiNews 2023 archive~\cite{Wikinews} (to ensure that the knowledge does not exist in LLMs). 
In terms of updating or erasing knowledge, we picked 200 questions from WebQuestions~\citep{berant2013webquestion} as knowledge that the language model can correctly answer without additional knowledge (to ensure that the knowledge exists in LLMs). Following this, all acquired knowledge is added to DynaMind's long-term memory. We employed GPT-4 to produce a question-and-answer pair for each unit of knowledge, and then prompted DynaMind to generate the corresponding answer. The answer accuracy of different models in DynaMind is recorded in Table~\ref{tab:knowledge_manipulation}. A higher score indicates a robust capacity for continuous assimilation of new knowledge for creation and update indicators, whereas a lower score is more desirable for deletion tasks. GPT-4 possesses advantages in both knowledge creation and update tasks. However, it is noteworthy to mention that Falcon has outperformed GPT-4 in the deletion task, yielding the most impressive results.

\begin{table}
\centering
\renewcommand\tabcolsep{6.2pt}
\begin{tabular}{lccc}
\toprule
\textbf{Model Name} & \textbf{Create}   & \textbf{Update}       & \textbf{Delete}   \\ \midrule
OpenAI GPT-3.5      & 92.0              & 81.0                  & 71.5              \\
OpenAI GPT-4        & \underline{95.5}  & \underline{85.0}      & 71.5              \\
Falcon-40B          & 90.5              & 83.5                  & \underline{74.0}  \\
Llama-30B           & 88.0              & 78.5                  & 70.0              \\ \bottomrule
\end{tabular}
\caption{Performance on Knowledge Manipulation.}
\vspace*{-1.5em}
\label{tab:knowledge_manipulation}
\end{table}

\vspace*{-0.5em}
\section{Conclusion}
\vspace*{-0.5em}
In this paper, we introduced DynaMind, a powerful continuous learning framework designed for large language models. By incorporating memory management and modular operators, DynaMind enables continuous integration of new knowledge into the cognitive process of these models, allowing them to adapt and improve over time. Through a series of benchmark evaluations, we have demonstrated the remarkable effectiveness of DynaMind, validating its ability to enhance the performance of large language models. 

% \clearpage

\bibliography{anthology,custom}
\bibliographystyle{acl_natbib}

\clearpage

\appendix
\onecolumn
\section*{Appendix}
\section{Formats and Examples of DynaMind Operators}
\begin{table}[h]
\centering
\resizebox{\textwidth}{!}{
\begin{tabular}{|l|l|l|}
\hline
\multirow{4}{*}{Coordinator}   & Format & Context: List{[}Knowledge: str{]}, Query: str → List(\textless{}Operator: str, Parameter: dict\textgreater{})                                                                                                                                                                                                                                                                                                                                                                                         \\ \cline{2-3} 
                               & Prompt & \begin{tabular}[c]{@{}l@{}}Context: \{Context\} Query: \{Query\}\\ You should only respond in JSON format, exclusively using the listed commands.\end{tabular}                                                                                                                                                                                                                                                                                                                                        \\ \cline{2-3} 
                               & Input  & \begin{tabular}[c]{@{}l@{}}\{"Query": "The 1995 Tooheys 1000 driver who has second-to-last \\ in the Tooheys Top 10 was born where?"\}\end{tabular}                                                                                                                                                                                                                                                                                                                                                   \\ \cline{2-3} 
                               & Output & \begin{tabular}[c]{@{}l@{}}{[} \textless{}"Searcher", \{"query": "The official record of the 1995 Tooheys 1000 race"\}\textgreater{},\\ \textless{}"Browse", \{"Path": "\textless{}The official record of the 1995 Tooheys 1000 race\textgreater{}", \\    "Query": "The top 10 drivers in the 1995 Tooheys 1000 race"\}\textgreater{},\\ \textless{}"Coordinator", \{"Query": "The birthplace of the driver who finished \\ second-to-last in the Tooheys Top 10"\}\textgreater {]}\end{tabular} \\ \hline
\multirow{4}{*}{Searcher}      & Format & Query: str → Results: List{[}Result: dict{]}                                                                                                                                                                                                                                                                                                                                                                                                                                                          \\ \cline{2-3} 
                               & Prompt & Search \{query\} in a search engine.                                                                                                                                                                                                                                                                                                                                                                                                                                                                  \\ \cline{2-3} 
                               & Input  & \{"Query": "The official record of the 1995 Tooheys 1000 race"\}                                                                                                                                                                                                                                                                                                                                                                                                                                      \\ \cline{2-3} 
                               & Output & \begin{tabular}[c]{@{}l@{}}{[}    \{       "desc": "The 1995 Tooheys 1000 was the 36th running of the Bathurst 1000\\       touring car race...", "url": "https://en.wikipedia.org/wiki/1995\_Tooheys\_1000/" \\    \} {]}\end{tabular}                                                                                                                                                                                                                                                         \\ \hline
\multirow{4}{*}{Browser}       & Format & Path: str, Query: str → Answer: str                                                                                                                                                                                                                                                                                                                                                                                                                                                                   \\ \cline{2-3} 
                               & Prompt & \begin{tabular}[c]{@{}l@{}}Given the \{Context\}, find the answer for \{Query\};\\ Return "None" if you cannot find the answer in the context.\end{tabular}                                                                                                                                                                                                                                                                                                                                           \\ \cline{2-3} 
                               & Input  & \begin{tabular}[c]{@{}l@{}}\{    "Path": "https://en.wikipedia.org/wiki/1995\_Tooheys\_1000",\\    "Query": "The top 10 list of drivers who participated in the 1995 Tooheys 1000 race" \}\end{tabular}                                                                                                                                                                                                                                                                                           \\ \cline{2-3} 
                               & Output & \textless{}The Top 10 list of drivers in the 1995 Tooheys 1000 race\textgreater{}                                                                                                                                                                                                                                                                                                                                                                                                                     \\ \hline
\multirow{4}{*}{Responder}     & Format & Context: List{[}Knowledge: str{]}, Query: str → Response: str                                                                                                                                                                                                                                                                                                                                                                                                                                         \\ \cline{2-3} 
                               & Prompt & Given the \{Context\} and the user’s query \{Query\}, generate the response to the user.                                                                                                                                                                                                                                                                                                                                                                                                              \\ \cline{2-3} 
                               & Input  & \begin{tabular}[c]{@{}l@{}}\{    "Context": {[}\textless{}Retrieved Knowledge\textgreater{}{]}, \\    "Query": "The 1995 Tooheys 1000 driver who has second-to-last \\    in the Tooheys Top 10 was born where?" \}\end{tabular}                                                                                                                                                                                                                                                                  \\ \cline{2-3} 
                               & Output & \begin{tabular}[c]{@{}l@{}}\{    "Response": "The second-to-last driver in the 1995 Tooheys 1000 race is \\     Tony Longhurst. He was born in Sydney." \}\end{tabular}                                                                                                                                                                                                                                                                                                                           \\ \hline
\multirow{4}{*}{Discriminator} & Format & Context: List{[}Knowledge: str{]}, Query: str, Response: str → Validity: boolean                                                                                                                                                                                                                                                                                                                                                                                                                      \\ \cline{2-3} 
                               & Prompt & Given the \{Context\}, check if the \{Response\} satisfies the \{query\}.                                                                                                                                                                                                                                                                                                                                                                                                                             \\ \cline{2-3} 
                               & Input  & \begin{tabular}[c]{@{}l@{}}\{    "Context": {[}\textless{}Retrieved Knowledge\textgreater{}{]}, \\    "Query": "The 1995 Tooheys 1000 driver who has second-to-last in the \\    Tooheys Top 10 was born where?","Response": "The second-to-last driver \\    in the 1995 Tooheys 1000 race is Tony Longhurst. He was born in Sydney." \}\end{tabular}                                                                                                                                            \\ \cline{2-3} 
                               & Output & True                                                                                                                                                                                                                                                                                                                                                                                                                                                                                                  \\ \hline
\end{tabular}
}
\caption{Formats and examples of DynaMind operators}
\label{tab:operators_details}
\end{table}

\newpage

\section{DynaMind Screenshot}
\begin{figure}[h]
    \centering
    \includegraphics[width=\textwidth]{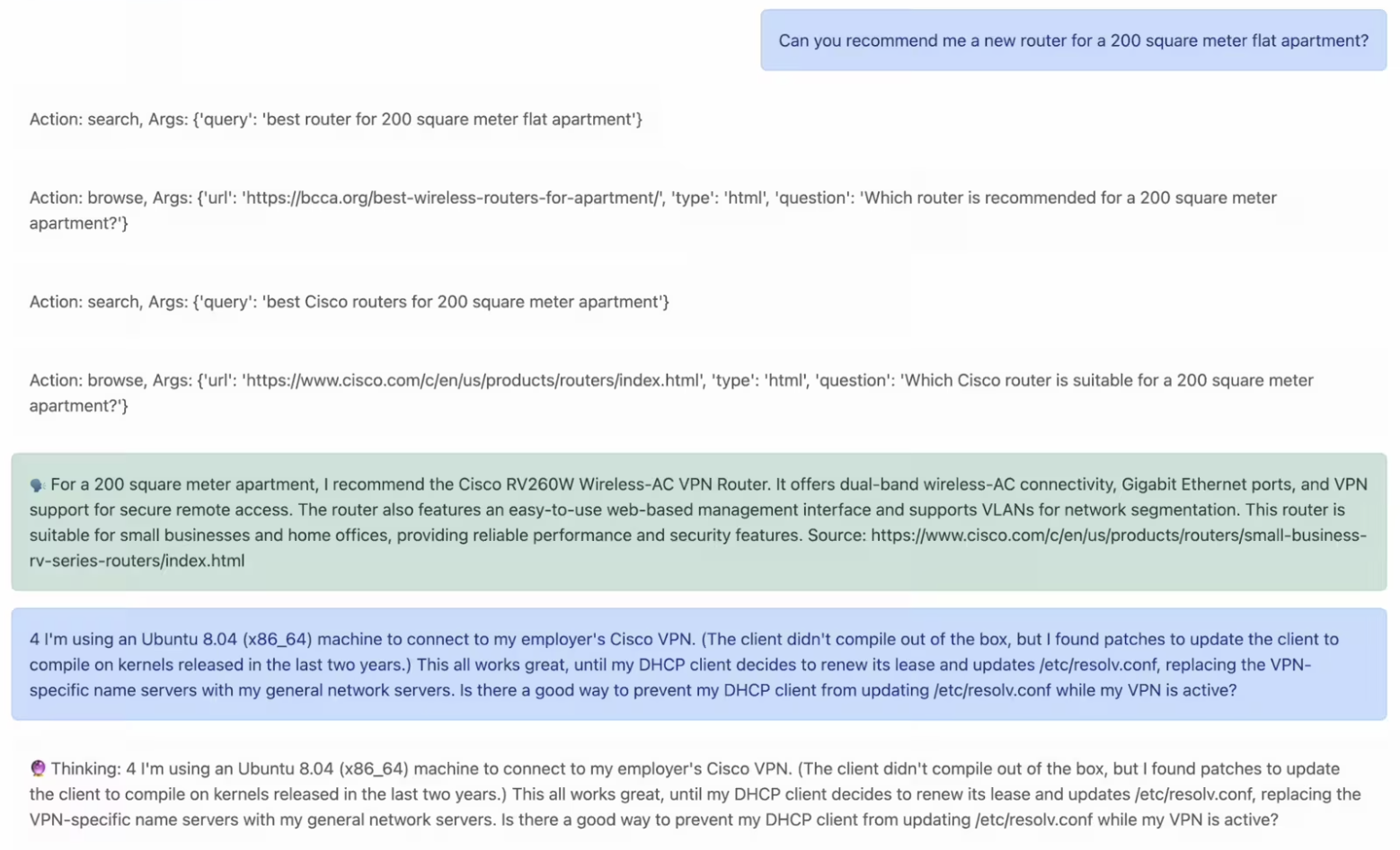}
    \caption{DynaMind screenshot.}
    \label{fig:dynamind_screenshot}
\end{figure}

\end{document}